\documentclass[sigconf,article]{acmart}

\renewcommand\footnotetextcopyrightpermission[1]{}
\usepackage{graphicx}
\graphicspath{{./}}
\usepackage{float}
\usepackage[capitalise,nameinlink]{cleveref}

\usepackage{acro}
\DeclareAcronym{pr}{short = PR, long = passage retrieval}
\DeclareAcronym{ac}{short = AC, long = answering with confidence}
\DeclareAcronym{pg}{short = PG, long = team-generated passages}
\DeclareAcronym{po}{short = PO, long = organiser-provided passages}
\DeclareAcronym{hmr}{short = HMR, long = harmonic mean of rewards}
\DeclareAcronym{auc}{short = AUC, long = area under the receiver operating characteristic curve}

\begin{document}

\title{BITEM at the NTCIR-19 R2C2 Task: Predicting Confidence from Agentic RAG Pipeline Signals}

\author{Julien Knafou, Luc Mottin, Alexandre Flament, Paul van Rijen, Esteban Gaillac,
        Patrick Ruch}
\affiliation[obeypunctuation=true]{
    \institution{HES-SO}, \city{Geneva}, \country{Switzerland}
}
\affiliation[obeypunctuation=true]{
    \institution{SIB Swiss Institute of Bioinformatics}, \city{Geneva}, \country{Switzerland}
}
\email{julien.knafou@hes-so.ch}

\begin{abstract}
The BITEM team entered both subtasks of the NTCIR-19 R2C2 task with a single agentic pipeline, in which a model searches, reads and records evidence over a movie corpus while an orchestrator holds the record and rules on what may be submitted. A claim is admitted only once an entailment cascade has checked it against the passage it cites, and an answer is released only once enough checked evidence stands behind it. Each question is run three or four times, every pass retrieving from a corpus stripped of what the earlier passes have already seen. The confidence score predicted with each answer is computed by the orchestrator from what the run leaves behind and is never asked of the model, which is offered no way to rate itself. The two retrieval runs placed 4th and 5th of 22, pooling the passes was worth 0.0709 nDCG@20, and the gain was largest on the multi-hop and post-processing-heavy questions, where the organisers rank the pooled run top of the field. Sixteen of the 25 answer runs were built on passages these two runs supplied, 12 of them submitted by other teams. HMR rewards a system whose confidence is high where it answers right and low where it answers wrong. The pipeline reached an accuracy of 0.9219, 6th of 25, while the confidence scores predicted for those answers gave an HMR of 0.4915, 13th. A few rules crafted over those same recorded signals, with no further model call and no further retrieval, raise that to an accuracy of 0.9375, 5th, and an HMR of 0.6985, 9th. Ranking on HMR alone can reward a system for answering wrongly with low confidence, so we propose accHMR, the accuracy multiplied by HMR, which reports the reward in proportion to the accuracy, and on which the revised rules would have scored 0.6549, 5th. For future work, fitting a model on the numbers the pipeline already produces, rather than writing such rules by hand, would be a real step forward.
\end{abstract}


\maketitle
\pagestyle{plain}
\acresetall

\section*{Team Name}
BITEM

\section*{Subtasks}
Passage Retrieval (PR) subtask (English)\\
Answering with Confidence (AC) subtask (English)

\section{Introduction}

For our participation in the NTCIR-19 R2C2 task~\cite{r2c2overview}, a system serving both subtasks was built around a single model. It holds eight tools, searches a corpus it has not seen, reads what it retrieves, and writes down the evidence it means to rely on. An orchestrator outside the model checks each piece of that evidence against the passage it cites with a two-model entailment cascade, and refuses to let the model submit an answer until enough evidence has passed that check. Once the answer is submitted, the confidence is computed by that same orchestrator over what the run has recorded and the model is offered no way to rate itself. One pass produces both a ranked list of passages and an answer with a confidence, so both subtasks are served by the same pipeline's artefacts. Each question is put through three or four such passes, a question stopping early when two passes in a row came back unsure. Passages a pass has seen are withheld from the next, so each searches a smaller corpus than the one before it. We submitted two runs for the \ac{pr} subtask and four for the \ac{ac} subtask.

Our two runs placed 4th and 5th of the 22 runs for the \ac{pr} subtask, 2nd of the seven teams that submitted, and the organisers' own test does not separate our better run from any run above it. While our system was among the most accurate in the \ac{ac} subtask, it failed to produce low confidence answers when it should have, which is what the \ac{hmr} rewards and what it scores independently of the accuracy. However, we found out that when adjusting the \ac{hmr} score to the accuracy, our best system's run ranks 7th instead of 13th out of 25 runs. Indeed, a look at \Cref{fig:acchmr} shows that half of the runs with high \ac{hmr} come with a low accuracy which sometimes can go down to around 50\%. It shows that the real challenge is to be able to climb that \ac{hmr} ladder while maintaining a high accuracy. This is discussed further in \Cref{sec:5}. Assembling the passes helped both subtasks, though not in the same way, for \ac{pr} it raised retrieval by 0.0709 nDCG@20, a difference the organisers' test does not establish, while for \ac{ac} it left the accuracy where it was and improved the confidence calibration instead.

After testing a few different ways of combining our system's passes, we found that a small set of common-sense rules, requiring only a few programming changes on the already computed artefacts, achieves an \ac{hmr} score close to 0.70 while also improving accuracy. Those rules read nothing the pipeline did not already produce, which shows that simple rules over a pipeline's own signals are enough to improve \ac{hmr}. These changes draw partly on the error analysis of \Cref{sec:4.2} and are described in \Cref{sec:signal}.

\section{Related Work}\label{sec:2}

Our system works each question over repeated turns, a model with eight tools searching a fixed corpus, reading passages and recording claims as evidence, while an orchestrator outside the model executes the calls, keeps the record and decides what may be submitted. Each question is run three or four times, each pass retrieving from a corpus stripped of everything its predecessors saw. \citet{li2025searcho1} give the loop its canonical form, a model that pauses its reasoning to search and resumes on what it finds, and \citet{zheng2025deepresearcher} train that behaviour against the live web. Such agents repeat themselves, 32\% of the steps near-duplicating the previous query in one study of multi-turn agent traffic~\cite{ning2026agentic}, and \citet{murali2026divinit} treat parallel threads converging on one retrieval path by diversifying the opening turn's queries.

In our system the model records a claim only when it has quoted a span from one passage, and an orchestrator outside the model then checks each recorded claim by entailment against the passage it cites, admitting only those that pass. An answer reaches submission only once enough admitted claims stand behind it, and the confidence score predicted for it is computed from those claims. \citet{honovich2022true} evaluate such checking across nine grounding datasets, where the best entailment model reaches an \ac{auc} of 81.5 against 63.8 for token-level F1. \citet{liu2023evaluating} check after the fact, auditing four generative search engines and finding 51.5\% of their statements fully supported by the citations given for them. \citet{qian2025vericite} verify a draft answer's statements before it is delivered, discarding the statements that fail.

Supervised methods learn the number from labelled data, buying their calibration with examples whose right answers are known in advance. \citet{kadavath2022know} fine-tune a model with an added value head, an extra output trained alongside the usual one, to predict whether it knows an answer before giving it. \citet{joren2025sufficient} judge from the question and the retrieved context, with no ground truth answer, whether that context is sufficient to answer, and fit that signal together with a self-rated answer probability in a logistic regression that predicts hallucination, thresholding its score to decide when to abstain.

Unsupervised methods take the number from a model as it stands, with nothing trained and nothing fitted. \citet{kadavath2022know} also report a training-free result in the same study, showing that a model given a handful of worked examples and no fine-tuning can judge whether the answer it just proposed is correct. A preference-tuned model can instead verbalise a confidence, stating it as a number in its own output. \citet{tian2023just} find that number typically better calibrated than the model's conditional probabilities, the probability it assigns to the answer itself. \citet{xiong2024express} benchmark such elicitation across five models and five kinds of task and find the verbalised numbers overconfident, clustered high, and only narrowly behind methods that read the model's own logits. Each of these estimates has a model rate its own output, which a confident error escapes, and \citet{kuhn2023semantic} avoid that rating by clustering the model's repeated answers into groups that mean the same thing and taking the entropy over those meanings rather than over the wordings, with each sample's own likelihood summed into the meaning it belongs to. \citet{vashurin2025benchmarking} group these routes into three families, information-based methods that read the model's logits and their entropy, sample-diversity methods that compare repeated generations, and reflexive methods that ask the model itself. Across 11 datasets they find which family works best to depend on the task, recommending information-based methods for short answers and sample diversity for longer generations.

\section{Method}

\begin{figure*}[t]
  \centering
  \includegraphics[width=0.86\textwidth]{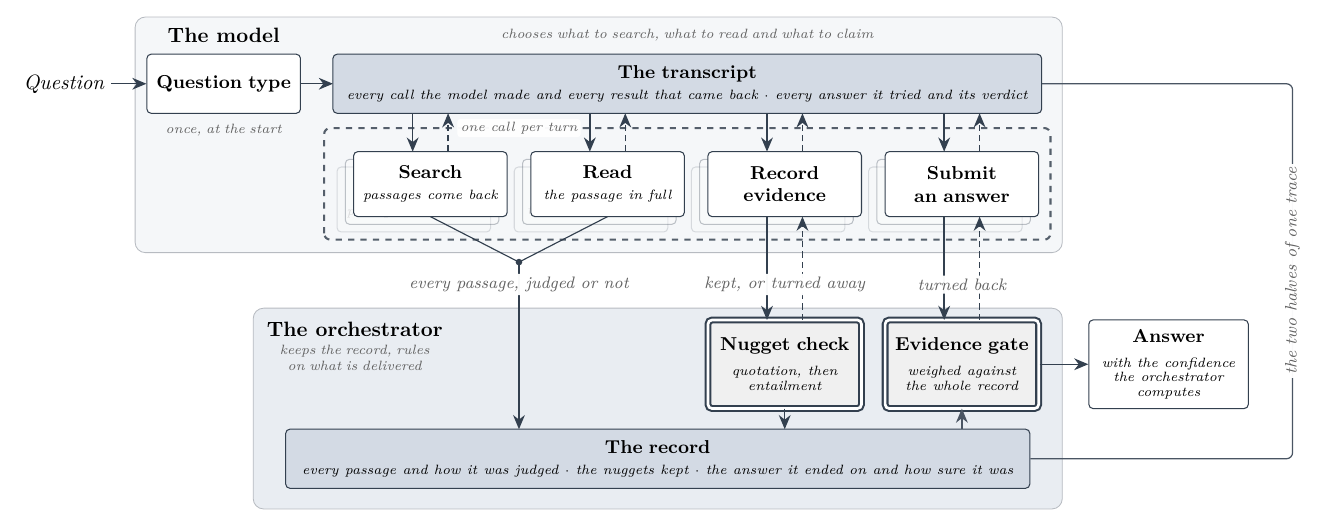}
	\Description{A diagram in two horizontal bands. The upper band, the model, holds the transcript above four boxes the model may call one per turn, drawn as a stack to show repetition, each box exchanging a call and a result with the transcript. The lower band, the orchestrator, holds two gates and the record. Arrows cross between the bands, solid downward for what the model proposes and dashed upward for what the orchestrator rules. A bracket down the right margin joins the transcript to the record as the two halves of one trace.} \caption{One question through the pipeline, as defined in \Cref{sec:3.3}}
  \label{fig:pipeline}
\end{figure*}

Similarly to the unsupervised methods of \Cref{sec:2}, our system has nothing trained or fitted for this task, and it differs from them in which signal is used to infer the confidence. Every confidence method there reads it from the model, from its logits, from repeated samples of its output or from what it says about itself, while ours is computed outside the model by the orchestrator, over a record of evidence that has been checked against the passages it was quoted from, and the model cannot give a confidence of its own. Our main contribution is this confidence score, more than the retrieval or the ensemble behind it. It is computed from signals any agentic pipeline already records, such as the number of searches a pass issued or the entailment scores of its evidence. \Cref{sec:4.2} evaluates the confidence of our official runs, and \Cref{sec:signal} shows that simple rules over the same signals improve it. Our predicted confidence comes from the distinct passages carrying a verified claim, so a question with little evidence receives a low confidence.

In what follows the language model reads, writes and chooses what to do next, and the orchestrator, the program outside the model, executes what the model asks for, keeps the record of what has happened, and rules on which answer is finally accepted.

\subsection{Preprocessing}\label{sec:3.1}

\subsubsection{Chunking}\label{sec:3.1.1}

We cut our own passages from the supplied wikitext rather than use the ones already cut for us, because a fixed window with overlap gives a statement two addresses. Our passages are aligned to sentence boundaries and packed greedily within a section to a 200-word target under a 300-word ceiling, with no overlap between neighbours. Each table becomes a passage of its own, never folded into anything. Every passage carries its article title, and one from a named section carries the header tree up to the leaf. The corpus we index holds 2,091,427 passages.

\subsubsection{Named entities}\label{sec:3.1.2}

Questions over this corpus turn on named entities, and the passage that answers a question often lacks the name the question uses, because the people, places and works an article discusses are usually named once and referred to obliquely thereafter. Annotating entities gives every passage the names it is about. Entities were extracted with one call per document using Qwen3.5-9B, which returns name-and-type lines over seven types, namely person, character, location, organisation, event, work and date, with character kept apart from person. An entity attaches to a passage when its name occurs as a substring of that passage's title, heading or text. That rule cannot reach the oblique mention it exists to solve, so a second call over each document asks which of the names the document carries belong to a passage that does not spell them out, and just under half the annotations we index arrive that way rather than by the substring rule.

\subsubsection{Indexing}\label{sec:3.1.3}

Retrieval uses two structures over the annotated passages. The first, the sparse index, serves the question that names its entity. It is an inverted index in Elasticsearch and the only one of the two that stores passage text, with the title, the section heading and the entity annotations held in fields of their own.

The second, the dense index, serves the question that describes an entity without naming it. Every passage is embedded once by nomic-embed-text-v1.5~\cite{nussbaum2024nomic} into a 768-dimensional vector. The text given to the encoder concatenates the first 10 of the passage's annotated names ahead of its title, section and contents.

A third mechanism, a knowledge graph of co-occurring names, was also built, but a normalisation mismatch left it matching only about one search in 60, and nothing it returned reached a submitted run. It would have been interesting to see how much the \ac{pr} would have improved had the mechanism been triggered as intended.

\subsection{Tools}\label{sec:3.2}

\subsubsection{Retrieval}\label{sec:3.2.1}

The model reaches passage text through \texttt{search\_\allowbreak{}passages}, in a call carrying one or more query lines, reformulations of the question. Every line goes to the inverted index and, encoded, to the dense index, each index keeping a passage's highest score across lines.

The two scores are not comparable, so reciprocal rank fusion~\cite{cormack2009reciprocal} merges the two rankings by position, lifting a passage by how highly and how often the searches rank it. The head of that merged list is then read by a cross-encoder, ms-marco-MiniLM-L-12-v2, which takes a query line and the passage together and so tells a passage sharing the question's vocabulary from one answering it~\cite{nogueira2019passage,reimers-2019-sentence-bert}. \texttt{search\_\allowbreak{}entities} matches a name against the annotated names and returns the names carried by the passages it hits, most frequent first, so the model can settle on the name the corpus uses.

\subsubsection{Reading}\label{sec:3.2.2}

Results are cut to a fixed length, so \texttt{get\_\allowbreak{}passage} returns named passages whole, and can add neighbouring passages on either side. Prose passages do not overlap, so reading a neighbour recovers a statement split across a cut.

\texttt{extract\_\allowbreak{}from\_\allowbreak{}passage} locates a sentence inside one passage. Given a short question in plain words, it scores the passage's own sentences against that question lexically, falling back to entailment scoring by MiniCheck-Flan-T5-Large~\cite{tang2024minicheck} where the passage words it differently, and returns the best few. This is how the model finds the words it will later have to quote.

\subsubsection{Recording evidence and submitting an answer}\label{sec:3.2.3}

Reading does not commit the model to anything. Evidence is recorded by \texttt{record\_\allowbreak{}evidence}, in a call that names a passage and may attach claims, each with the span quoted from that passage. The orchestrator keeps a claim with its passage as a nugget, adding each further passage on which the model records it.

The quotation check requires the span to appear in the passage verbatim or nearly so, and the entailment check then asks whether the span really says what the claim says, reading the pair with MiniCheck-Flan-T5-Large and a DeBERTa-v3-large entailment model~\cite{he2023debertav3}. \texttt{list\_\allowbreak{}nuggets} lists what has been recorded, and \texttt{flag\_\allowbreak{}nugget\_\allowbreak{}bogus} was available to retract a nugget that later reading contradicts, though no pass used it.

The answer is submitted through \texttt{submit\_\allowbreak{}answer}, with a rationale and the passages that ground it. When nothing supports an answer the model raises a refusal flag instead and leaves the wording to the orchestrator. The orchestrator rules on it, ending the pass or returning the model to the loop.

\subsection{Pipeline}\label{sec:3.3}

\subsubsection{The question}\label{sec:3.3.1}

The task draws its questions from six question types~\cite{r2c2design,yang2024crag}, and none arrives with its type label, so Qwen3.5-122B-A10B-FP8, the model that goes on to answer it, is asked to assign one. The type rides into the opening message as advice on what form the answer should take.

The tools of \Cref{sec:3.2} are attached from that message onward, and everything the model then proposes through them crosses to the orchestrator to be recorded or turned back, the shape \Cref{fig:pipeline} draws. The record is the pass's own account of its work, separate from the transcript the model reads, and holds the passages the pass retrieved under the status each was given, the nuggets recorded against them with their entailment scores, the answer reached and its confidence, and how many searches were issued and how often an answer was turned back.

\subsubsection{The loop}\label{sec:3.3.2}

The model works one turn at a time, emitting a tool call the orchestrator executes before the next turn opens. Every reply carries a running count of the tool calls spent against the budget advertised to the model, which is advice rather than a limit.

The orchestrator refuses a \texttt{record\_\allowbreak{}evidence} call naming a passage the record does not hold, so the model can cite only what search and reading have put in front of it. The orchestrator also writes to the record on its own terms.

The sequence ends when the orchestrator accepts an answer, when a reply carries no tool call at all, in which case the orchestrator keeps its prose as a candidate answer, or when the transcript outgrows the context window.

\subsubsection{Committing an answer}\label{sec:3.3.3}

An answer submitted through \texttt{submit\_\allowbreak{}answer} must first satisfy the organisers' file format, i.e. an answer and its confidence sharing one semicolon-delimited line. The orchestrator then counts the distinct passages that verified nuggets rest on, so a claim recorded on two passages contributes two. An answer needs two such passages, or one entailing its claim very strongly, and a confidence above a floor. Only that confidence sets the answer against the model's recorded claims, each verified claim weighted by how much of the answer's wording appears in that claim, by word overlap, never by the entailment cascade. Because that weight never vanishes, the overlap can move the confidence but never drops a claim from the count. Each refusal tells the model what it holds and asks it to try again, and the attempt remains in the transcript with the answer the model wrote and the confidence computed for it. After several refusals the fallbacks take over, one of which \Cref{sec:3.4} describes.

\subsubsection{Confidence}\label{sec:3.3.4}

An accepted answer leaves with a confidence the orchestrator computes over the record, and never one the model proposes, since \texttt{submit\_\allowbreak{}answer} takes no confidence argument. Where the fallbacks delivered the answer instead, the confidence is fixed by which fallback fired. How each official run turns those numbers into its final confidence differs from run to run, and \Cref{sec:3.5} gives each rule. The confidence is estimated outside the model, from what the run recorded and not from what the model says about itself. Computing it requires no second model, no further retrieval and no change to the loop, so the estimator can be separated from the system it scores. It can use any quantity an agentic pipeline already records, such as how many searches a pass issued, how often it put an answer up and was turned back, what the entailment cascade scored the evidence it cited, and how long its longest turn of reasoning ran.

\subsection{Ensemble strategy}\label{sec:3.4}

In the first pass on a question the retrieval tool searches the whole corpus. Each later pass searches a corpus of passages with the ones its predecessors recorded removed. This strategy serves two goals, retrieving more important passages for \ac{pr} and building the confidence score by comparing the answers across passes. Three passes are run on every question, and a fourth unless the last two both came back unsure, so the questions the system is least certain of are the ones it attempts fewest times. The passes are not independent draws, because where a pass cannot reach an answer of its own, one of the fallbacks offers it the most confident answer an earlier pass on the same question committed and adopts that answer if a passage entails it, so passes that agree may have seen each other's answers.

\subsection{Runs}\label{sec:3.5}

A pass writes one trace when it ends, and the trace holds both the transcript and the record. The transcript is the conversation, every tool call the model made and every result the orchestrator returned, so every answer the model submitted survives there with the verdict passed on it, and a submission the orchestrator refused keeps both the wording the model wrote and the confidence computed for it. The record is the orchestrator's account of the pass, holding the passages it retrieved with the status each was given, the nuggets with their entailment scores, the answer with its confidence, and counts of what the pass did, such as the searches it issued and the answers it had turned back. The runs are built from those traces, a passage retrieval run drawing on the record alone, an answer run on the record for the nuggets it includes and on the transcript for the attempts it chooses among, since the record keeps only the answer a pass ended with.

\subsubsection{Passage retrieval}\label{sec:3.5.1}

A \ac{pr} run gives, for each question, a ranked list of at most 20 passages, each with its text and the document it was cut from.

\textbf{BITEM-PG-1}, the \emph{single pass} run, is built from the first pass on each question. The passages that pass kept evidence on come first, ordered by how strongly the quoted span entailed the claim made against it, and the rest follow in the order the pass met them.

\textbf{BITEM-PG-2}, \emph{pooled}, is built from all the recorded passes on a question. Different passes can reach different answers, so they are first grouped by the answer they reached, and the group whose strongest pass was most confident, and which the most passes joined, supplies the run. Every passage every pass in that group retrieved is pooled, not only those its nuggets rest on, ordered by entailment as before.

\subsubsection{Answering with confidence}\label{sec:3.5.2}

An \ac{ac} run gives, for each question, an answer and a confidence, followed by the nuggets supporting it, each naming a passage retrieval run and a rank within it.

\textbf{BITEM-AC-1}, the \emph{single pass} run, is built from the first pass on each question and paired with BITEM-PG-1. It takes the answer that pass committed and the confidence that came with it, already raised by a term meant to reward agreement across passes, which with a single pass is a fixed fraction of the distance to 100. Where the pass committed a refusal, the run returns to the transcript, takes every answer that pass submitted, discards the refusals among them, and keeps the one that got furthest through the gates of \Cref{sec:3.3.3}, ties broken by the confidence the orchestrator computed for each submission, keeping whichever is higher of that confidence and the committed one. The nuggets are the verified ones in that pass's record, and a nugget is kept, citing one of its own passages that appears in BITEM-PG-1, or dropped when none does.

\textbf{BITEM-AC-2}, \emph{most confident}, and \textbf{BITEM-AC-3}, \emph{summed agreement}, are built from all the recorded passes and paired with BITEM-PG-2. Both pool from the transcripts every answer submitted through \texttt{submit\_\allowbreak{}answer}, including those the orchestrator turned back, and drop only the answers worded as refusals. BITEM-AC-2 takes the answer of the one submission whose confidence was highest and keeps that confidence unchanged. BITEM-AC-3 groups the submissions whose answers match after normalisation, and selects the wording of the group whose confidences sum highest, with that sum capped at 100. Their nuggets come from the records of those passes rather than the transcripts, and are matched to BITEM-PG-2 as BITEM-AC-1's are to BITEM-PG-1. That matching normally looks at neither the answer nor the passages the chosen submission cited, so the two runs cite the same nuggets on all 65 questions. A fix to a single question the day before the deadline rebuilt our fourth file from BITEM-AC-3 rather than from its own strategy, so BITEM-AC-3 was submitted twice and scored twice. The model as a judge approach is described and evaluated in \Cref{sec:judge}.

\section{Results}\label{sec:4}

We answered 65 questions and the official evaluation scores 64, question 0025 having been reissued after release. In both subtasks the comparison our runs were built for is the one of \Cref{sec:3.5}, a single first pass against the recorded passes assembled.

\subsection{Passage retrieval}\label{sec:4.1}

\begin{table}[t]
\centering
\caption{Passage retrieval under the consolidated direct qrels, over the 64 scored questions. $\Delta$ is the mean per-question difference, pooled minus single pass. W/T/L counts the questions on which the pooled run scores higher than, the same as, and lower than the single-pass run. The better run is bold. The organisers' own randomised Tukey HSD does not separate the two runs, at $p = 0.4854$ on nDCG@20 and $p = 0.1086$ on Q@20~\cite{r2c2overview}.}
\label{tab:pr}
\footnotesize\setlength{\tabcolsep}{3.5pt}
\begin{tabular}{lcccc}
\toprule
 & \multicolumn{1}{c}{PG-1} & \multicolumn{1}{c}{PG-2} & & \multicolumn{1}{c}{PG-2} \\
 & \multicolumn{1}{c}{\itshape single pass} & \multicolumn{1}{c}{\itshape pooled} & $\Delta$ & \multicolumn{1}{c}{W/T/L} \\
\midrule
nDCG@20 & 0.3509 & \textbf{0.4217} & $+$0.0709 & 42/7/15 \\
Q@20      & 0.2677 & \textbf{0.3582} & $+$0.0905 & 41/7/16 \\
\bottomrule
\end{tabular}
\end{table}

The task provides two sets of relevance judgements. We report only the direct one, in which assessors judged each passage against the question. The other was derived from the nuggets the answer runs cited, which points it at whichever retrieval runs those teams chose to answer from, and 12 of the 25 answer runs were built on ours. The organisers designate the direct judgements as the official evaluation and note that bias themselves~\cite{r2c2overview}.

As we can see in \Cref{tab:pr}, assembling the passes seems to improve retrieval. Indeed, the pooled run gains 0.0709 nDCG@20 and 0.0905 Q@20 over the single-pass run, winning 42 of the 64 questions against 15. Even though the pooled run scores higher than the single-pass run, the difference is not statistically significant, so the gain is an observed difference rather than an established one.

\begin{table}[t]
\centering
\caption{Mean nDCG@20 by question type under the direct qrels, in the organisers' order. $\Delta$ is pooled minus single pass, and W/T/L counts the questions the pooled run wins, ties and loses. The better run is bold.}
\label{tab:bytype}
\footnotesize\setlength{\tabcolsep}{4.5pt}
\begin{tabular}{lrcccc}
\toprule
 & & & & & \multicolumn{1}{c}{PG-2} \\
question type & $n$ & PG-1 & PG-2 & $\Delta$ & \multicolumn{1}{c}{W/T/L} \\
\midrule
simple                & 26 & 0.3034 & \textbf{0.3608} & $+$0.0574 & 14/6/6 \\
simple w/ condition   &  7 & 0.2888 & \textbf{0.3403} & $+$0.0515 & 6/0/1 \\
comparison            &  7 & 0.4168 & \textbf{0.4392} & $+$0.0223 & 5/0/2 \\
aggregation           &  7 & 0.4498 & \textbf{0.4859} & $+$0.0361 & 4/0/3 \\
multi-hop             & 10 & 0.3749 & \textbf{0.4747} & $+$0.0998 & 7/0/3 \\
post-processing-heavy &  7 & 0.3900 & \textbf{0.5725} & $+$0.1824 & 6/1/0 \\
\bottomrule
\end{tabular}
\end{table}

\Cref{tab:bytype} splits the gain by question type. Post-processing-heavy and multi-hop gain 0.1824 and 0.0998 against 0.0223 for comparison, and on exactly those two types the organisers report our pooled run as the top run of the whole field under both measures~\cite{r2c2overview}. These are the two types whose answers cannot be found in a single passage, such as a date computed across editions or a fact reached through an intermediate entity. They are also where withholding what an earlier pass has already seen forces the next onto articles it would otherwise never rank. The pooled run holds 5.3 more relevant passages per question on post-processing-heavy and 3.1 more on multi-hop, against 0.4 on comparison. Because a passage is one slice of an article, the pooled run also reaches more articles, 13.1 per question in its 20 slots against the single pass's 8.6, which for the same list length is fewer repeated visits to the same article. The passages the later passes add match the quality of those already in the list, being judged highly relevant at 15\%, the same rate as the single-pass list.

The gain in \Cref{tab:pr} is a per-question mean over the 64 scored questions, and it carries a caveat, which is why we rest the argument on the per-type ordering of \Cref{tab:bytype} instead. The single-pass run failed to fill its 20 slots on 37 of the 64 questions, 21 of them stopping at exactly 10, so the ensemble was often adding to a list with room rather than displacing anything. Cutting the pooled list back to the single-pass length halves the overall advantage to 0.0411. On the 37 questions where the first pass fell short the gain is 0.1104, and on the 27 where it returned a full list it is 0.0167. Ten is the search tool's default, and the model left it in place on 2,126 of its 2,329 searches. On each of those 21 questions the first pass searched once, was given the 10 results it asked for, and stopped, since the loop is told to submit as soon as two nuggets support an answer. How many passages a run returns is settled by how much searching the answering happened to need, so the short lists follow from that stopping behaviour and are not evidence that the corpus had nothing further to return. No significance test is reported here, since the organisers note that every type holds too few questions for these scores to be read as more than descriptive trends~\cite{r2c2overview}.

\subsection{Answering with confidence}\label{sec:4.2}

\begin{figure}[t]
  \centering
  \includegraphics[width=\columnwidth]{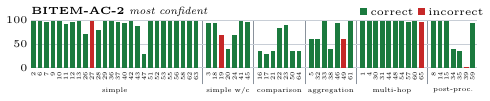}
\Description{A bar chart over the 64 scored questions. Each bar is the confidence BITEM-AC-2 predicted for that question, green where its answer was judged correct and red where it was not.}
  \caption{Confidence predicted by \emph{most confident} on each of the 64 scored questions, green where the answer was judged correct and red where it was not, and bars at zero drawn as a stub so their colour shows.}
  \label{fig:confidence}
\end{figure}

Assembling the passes did not make our answers more accurate. \Cref{tab:ac} reports the four runs we submitted, of which three are distinct systems, and their accuracy varies little, standing at 0.9062 for the single pass, 0.9219 for \emph{most confident} and 0.8906 for \emph{summed agreement}, with the same 56 of the 64 questions answered correctly in all three. With so few errors, the experiment could not show a difference in accuracy between the ways of assembling the passes. The confidence measures do separate the runs. The HMR is 0.2719 for the single pass, 0.4915 for \emph{most confident} and 0.3956 for \emph{summed agreement}, and the accHMR, the accuracy multiplied by HMR, is 0.2464, 0.4531 and 0.3523 respectively. Among the 25 answer runs, the single pass ranks 7th on accuracy, 19th on HMR and 19th on accHMR, \emph{most confident} 6th, 13th and 7th, and \emph{summed agreement} 8th, 16th and 14th. Of the 12 runs ranked above \emph{most confident} on HMR, six answered fewer than 70\% of the questions correctly. The single-pass confidence carries no information about whether its answer is right. Its \ac{auc}, the probability that a correct answer gets a higher confidence than a wrong one, is 0.463, below chance (0.5), and 12 of its 13 answers with a confidence below 50 were correct.

\begin{table}[t]
\centering
\caption{Our three official runs. The organisers define acc, $R_O$, $R_U$, HMR and avconf~\cite{r2c2overview}. accHMR is accuracy times HMR. Bold marks the best of the three.}
\label{tab:ac}
\scriptsize\setlength{\tabcolsep}{2.2pt}
\begin{tabular}{lc|cccc|c}
\toprule
 & acc & $R_O$ & $R_U$ & HMR & avconf & accHMR \\
\midrule
AC-1 \itshape single pass & 0.9062 & 0.1633 & 0.8102 & 0.2719 & 0.8127 & 0.2464 \\
AC-2 \itshape most confident & \textbf{0.9219} & \textbf{0.3460} & 0.8483 & \textbf{0.4915} & 0.8331 & \textbf{0.4531} \\
AC-3 \itshape summed agreement & 0.8906 & 0.2471 & \textbf{0.9904} & 0.3956 & 0.9644 & 0.3523 \\
\bottomrule
\end{tabular}
\end{table}

The errors of \emph{most confident} are few enough to read individually. They are the five red bars of \Cref{fig:confidence}. Three of the five are answered correctly by another team's run that used our two passage retrieval runs as its only retrieval input, so on those three the passages needed to answer were retrieved, and the error was made when the answer was written. The recurring defect is the answer string, which is either a fragment of the gold answer or a count that our own citations contradict. For example, asked what food Hannibal Lecter is associated with, \emph{most confident} answered \emph{fava beans} at a confidence of 100 where the gold is \emph{Liver with Chianti}, although its own nuggets quote the full sentence, \emph{I ate his liver with some fava beans and a nice Chianti}.

A further error was discovered in the aggregator rules, on a question where two passes reached different answers, each at 100, and the outcome between them followed from those rules. \emph{Summed agreement} adds every submission an answer attracted, and the wrong answer had been put to the evidence gate twice, once refused at 77.1 for insufficient evidence and once accepted at 100, which sums to 177.1 against the right answer's 100.0. The refused attempt supplies the whole margin, because a submission's weight is added whether or not the gate accepted it, so the rejected submission still counted toward the answer it had been rejected for. Counting one submission per pass leaves a tie, which the entailment summed behind each answer breaks in favour of the right one.

Counts are a second major source of error. Of the 64 questions, 15 ask how many or how old, and on four of them the passes propose different numbers, 5 against 8, 8 against 18, 8604 against 8612, and 1 against 2. Each pass counts over the nuggets it happened to record, and the passes recorded different subsets. Every nugget behind these answers is true, and none of them establishes that the list is complete (see \Cref{sec:3.3.3}), so nothing in the record says which pass read the whole set. On all four questions one of the passes proposed the right number, and it was not the one selected. On two questions the right answer was rejected for exceeding the length limit and replaced with an earlier pass's, and on two more it was reached by a pass the evidence gate had refused, force-delivered at the confidence floor, and outvoted by a pass that had committed cleanly and counted low.

\section{Additional experiments}\label{sec:rebuilt}

\begin{table}[t]
\centering
\caption{Results for model as a judge and two runs rebuilt by rule from the pipeline's own signals. accHMR is accuracy times HMR. Bold marks the best, ties included.}
\label{tab:revised}
\scriptsize\setlength{\tabcolsep}{2.8pt}
\begin{tabular}{lc|cccc|c}
\toprule
 & acc & $R_O$ & $R_U$ & HMR & avconf & accHMR \\
\midrule
\itshape model as judge (recovered run) & \textbf{0.9375} & 0.1750 & \textbf{0.9430} & 0.2952 & 0.9356 & 0.2768 \\
\midrule[\heavyrulewidth]
\itshape single pass, longest deliberation & 0.9062 & \textbf{0.8333} & 0.5517 & 0.6639 & 0.5156 & 0.6017 \\
\itshape four passes, independent agreement & \textbf{0.9375} & 0.6250 & 0.7917 & \textbf{0.6985} & 0.7656 & \textbf{0.6549} \\
\bottomrule
\end{tabular}
\end{table}

This section presents the recovered model as judge run and two other strategies we built from the signal. The organisers never judged these runs, so each of the 192 answers, three runs over 64 questions, takes the label of an identical answer they did judge, once lowercased with punctuation collapsed. For 175 of the 192, that answer is one our own official runs gave, since these strategies mostly reuse answers the pipeline had already produced. For 2 more, it is an answer only another team submitted. The remaining 15 had no judged match and were labelled by hand, beside other teams' answers to the same question. On the Hannibal Lecter question of \Cref{sec:4.2}, the recovered run answered \emph{liver with fava beans and a nice Chianti}, which no judged run had given. The gold, \emph{Liver with Chianti}, names two things, and answers naming either alone were marked right. Naming the beans made no answer wrong, since eight that name them were marked right, while all ten that stopped at the beans were wrong. Since our answer names both elements of the gold answer and adds only non-essential detail, we labelled it correct. The other 14 were labelled the same way. Of these, 13 were labelled correct, and the answer \emph{8} to a question whose gold is \emph{18} was labelled wrong.

\subsection{Recovered model as judge run}\label{sec:judge}

Model as judge is built from all the recorded passes and paired with BITEM-PG-2. It puts to the answering model every submission those passes made, with the confidence and the citations each carried, the verified nuggets behind them and the passages that run retrieved for the question, and requires one \texttt{submit\_\allowbreak{}answer} call giving an answer, a confidence and the passages it rests on, retrying when it cites one that run did not retrieve. The confidence is predicted by the model itself rather than by the orchestrator, which is true of no other run of ours.

In \Cref{tab:revised}, model as judge reaches an accuracy of 0.9375, above all our official runs, and an $R_U$ of 0.9430. Like our official runs, however, it kept a high confidence on its wrong answers, so its $R_O$ is only 0.1750, its HMR 0.2952 and its accHMR 0.2768.

\subsection{From signal to confidence}\label{sec:signal}

Building on the error analysis of \Cref{sec:4.2} and on further observations of the traces, we modified the last part of the aggregator of our runs, in an effort to show that with common-sense rules there is already enough signal in the trace to improve HMR. Both revised runs take as the answer the last string the model itself put to \verb|submit_answer| rather than what the format check made of it, which on three questions had replaced a correct answer with a refusal. Neither calls a model or issues a retrieval. A single pass leaves one trace, while four passes leave four that can be compared.

For the revised single pass, the rule can only decide how far to trust the one pass it has. The confidence computed by the orchestrator cannot be used for that, because it counts the passages that passed the evidence gate, so it measures how much was read and not how well the evidence supports the answer. The transcript, however, records how hard the pass worked. We take the length, in characters, of the pass's longest reasoning turn as a signal of the difficulty of the question, and assign a confidence of 100 when that turn is shorter than the median over the first passes of all questions, and 0 otherwise. Taken alone, over the first pass of each of the 64 questions, the longest turn ranks wrong answers above right ones with an \ac{auc} of 0.707, against 0.559 for the number of distinct queries and 0.586 for the number of tool calls, each also taken alone, so it does not simply measure how much work was done. In \Cref{tab:revised}, the single-pass run, based on this one signal only, reaches an accuracy of 0.9062, an HMR of 0.6639 and an accHMR of 0.6017.

For the revised four passes, each pass on a question contributes one proposal, the last answer it submitted through \texttt{submit\_\allowbreak{}answer} that is not a refusal, whether or not the evidence gate accepted it, so a submission the gate refused no longer adds weight to the answer it proposed. Proposals whose answers are identical once lowercased and stripped of punctuation form a group. Each pass in a group adds one vote to it, or 0.35 of a vote when the pass did not reach the answer itself but was handed an earlier pass's answer by a fallback (\Cref{sec:3.4}), and the group with the most votes is selected. Where two groups have equal votes, we choose the group whose best pass has the highest entailment score summed over its verified nuggets. The confidence we assign is 100 when at least two passes of the selected group reached its answer on their own, 50 when at least two passes made a proposal but no answer was reached on their own by two of them, and 0 when only one pass made a proposal. This run reaches an accuracy of 0.9375, an HMR of 0.6985 and an accHMR of 0.6549 (\Cref{tab:revised}).

Both revised runs therefore beat all our official runs on HMR and accHMR, and the four-pass run also beats the single-pass run on accuracy and HMR. Among the 25 official runs, the single-pass and the four-pass runs would have ranked 11th and 9th on HMR, and 7th and 5th on accHMR, respectively.

\section{Discussion and Conclusions}\label{sec:5}

Assembling several passes of one system helped both halves of this task. In retrieval it helped directly, though close to half of the gain is list length rather than better material. In answering it left the accuracy where it was and improved the confidence instead, and rules applied to the passes the system had already recorded raise accHMR from 0.4531 to 0.6549 without a further model call or a further retrieval. The passes those rules read were not free, and running four costs about four times running one. Whether that is well spent is best asked against a larger model at matched cost, since repeating the same model is linear in the number of passes while making the model larger is not. These rules were designed on the same 64 questions they are evaluated on, so they should next be tested on a question set we have never seen.

The rules we used were written by hand, and a natural next step would be to fit them instead. Every quantity they read is a number the pipeline already produces, so a regression could take its inputs from the trace as it stands, such as how many queries a pass issued, how much entailment stood behind what it cited, and how long it deliberated. Fitted on a separate collection, or on synthetic questions generated for the purpose, such a regression would predict a confidence at the end of a run without calling a second language model and without touching the pipeline that produced it. The R2C2 test set would then be held-out data on which the fitted estimator could be evaluated and compared with the hand-written rules of \Cref{sec:signal} and the confidence of our official runs, over the same questions and under the same measure.
\begin{figure}[tb]
  \centering
  \includegraphics[width=0.74\columnwidth]{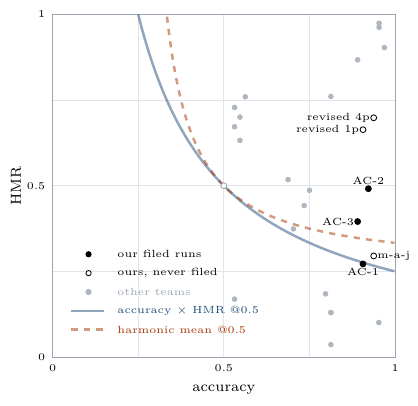}
\Description{A square scatter plot with accuracy on the horizontal axis and HMR on the vertical. Twenty-five grey points spread across the square. Three solid black points sit at high accuracy and low HMR, three hollow ones above them. Two curves cross at the centre, one for accuracy times HMR and one for a harmonic mean, both joining the pairs that rule scores as it scores a run at 0.5 on both.}
  \caption{Accuracy against HMR for every answer run submitted to the task, ours in black and the ones we never submitted hollow. Each curve joins the pairs its rule scores as it scores a run at 0.5 on both.}
  \label{fig:acchmr}
\end{figure}

HMR is great, because it asks a system to know when it is wrong and that is rarely evaluated at all. Our suggestion for a next iteration is that the headline ranking carry the accuracy alongside it. The two are scored independently by design, and \Cref{fig:acchmr} shows what that independence permits, with several runs reaching a high HMR at an accuracy near one half. A run predicting a confidence of 50 on every question scores exactly 0.5000 whatever it answers, as long as it answers at least one question wrongly and at least one correctly. Even worse, a run with an accuracy of 0.0000 predicting a systematic confidence of 0 would score an HMR of 1.0000. The accHMR metric would adjust it proportionally to the accuracy, meaning to 0.0000, which seems rather fair. The multiplication of both metrics was chosen as its interpretation is straightforward, the system's HMR gets validated proportionally to its accuracy. While other ways of aggregating both could have been considered, we did not think that it deserved much attention, HMR being an interesting metric, using it with a simple multiplication felt right and easy to process in a table next to both metrics.

Additionally, including questions that nothing in the corpus supports would help build such a measure. It would establish which questions a system ought to be able to answer, so that a system holding all the evidence and failing on half of them is penalised for it, and it would let a system earn a low confidence honestly, by reporting that it looked and found nothing rather than that it doubts what it found. It would also keep the measure loaded. HMR is the harmonic mean of $R_U$ and $R_O$, and $R_O$ is averaged over the wrong answers alone, so a system that answers every question correctly has nothing left to be overconfident about and takes an $R_O$ of one by default. Where some questions cannot be answered from the corpus, no system reaches that position for free, and a low confidence and a high accuracy become things to be earned together rather than traded against each other.

\bibliographystyle{ACM-Reference-Format}
\bibliography{r2c2}

\end{document}